# Interplay of ISMS and AIMS in context of the EU AI Act

Jordan Pötsch

*Abstract*— The EU AI Act (AIA) mandates the implementation of a risk management system (RMS) and a quality management system (QMS) for high-risk AI systems. The ISO/IEC 42001 standard provides a foundation for fulfilling these requirements but does not cover all EU-specific regulatory stipulations. To enhance the implementation of the AIA in Germany, the Federal Office for Information Security (BSI) could introduce the national standard BSI 200-5, which specifies AIA requirements and integrates existing ISMS standards, such as ISO/IEC 27001. This paper examines the interfaces between an information security management system (ISMS) and an AI management system (AIMS), demonstrating that incorporating existing ISMS controls with specific AI extensions presents an effective strategy for complying with Article 15 of the AIA. Four new AI modules are introduced, proposed for inclusion in the BSI IT Grundschutz framework to comprehensively ensure the security of AI systems. Additionally, an approach for adapting BSI's qualification and certification systems is outlined to ensure that expertise in secure AI handling is continuously developed. Finally, the paper discusses how the BSI could bridge international standards and the specific requirements of the AIA through the nationalization of ISO/IEC 42001, creating synergies and bolstering the competitiveness of the German AI landscape.

*Keywords*— EU AI Act, AI Governance, AI Management System, ISO/IEC 42001, AI Security, AI Standards

## I. Introduction

The EU AI Act (AIA) mandates, among other things, the implementation of a risk management system (RMS) (Article 1 + Article 8 + Article 9) and quality management systems (QMS) (Article 1, Article 16, Article 17, Annex VII) for high-risk AI systems [1, 2].

### A. ISO/IEC 42001 and the EU AI Act

To implement the AIA requirements, standards are necessary [3]. The ISO/IEC 42001 standard, published in December 2023, which describes an AI management system (AIMS), includes essential elements for fulfilling the aforementioned AIA articles [4]. However, the report by Soler Garrido et al. [5] shows that while ISO/IEC 42001, ISO/IEC 42005, and ISO/IEC 42006 are globally recognized standards, they do not directly address specific EU regulatory requirements mandated by the AIA. Soler Garrido et al. [6] discuss harmonized standards for the AIA and highlight the challenges involved.

The advantage of ISO/IEC 42001 lies in its High Level Structure (HLS), which facilitates the integration of a management system with minimal effort [3]. It builds on existing management systems, such as ISO/IEC 9001 and ISO/IEC 27001 [7]. To create synergies in practice, it is practical to build on existing risk processes within an organization when implementing an RMS for AI as required by the AIA [8]. The ISMS can serve as a foundation and starting point for integrating additional dimensions of trustworthy AI as part of the RMS [9]. This approach differentiates between existing ISMS controls that address AI security risks, supplements to these controls, and new controls needed to address novel security risks introduced by AI [8]. Subsequently, an AIMS based on ISO/IEC 42001 can be implemented with minimal effort, fulfilling further dimensions of trustworthy AI [8].

### B. Cybersecurity in the AI Act

The report by Soler Garrido et al. [5] states that an ISMS based on ISO/IEC 27001 is regarded as a fundamental basis for meeting the requirements of Article 15 of the AIA. However, the following challenges must be considered:

1. **Integration of management standards**: ISO/IEC 27001 needs to be aligned with other standards, such as ISO/IEC 42001, and future European standards on AI risk management.

2. **Expansion of AI-specific cybersecurity aspects**: Current standards lack emphasis on AI-specific cybersecurity risks.

3. **Inclusion of international standardization efforts**: Emerging standards like ISO/IEC 27090 provide guidance on AI cybersecurity. European contributions should align with the AIA to address specific AI-related risks.

Nolte et al. [10] conclude that the AIA leaves open the question of whether organizational measures under Article 15(5)(ii) and (iii) AIA are mandatory. Junkelwitz et al. [11] believe that both organizational and technical solutions must be implemented to meet the cybersecurity objectives of Article 15 AIA. ISO/IEC 27001 should serve as a foundation, with the addition of AI-specific security controls. ENISA's framework for AI good cybersecurity practices (FAICP) [12], which extends ISMS controls with AI-specific elements, is presented as an appropriate approach for complying with Article 15 AIA. Soler Garrido et al. [13] analyze IEEE standards in relation to AIA requirements and find that IEEE P2841 most closely meets these requirements. Kalodanis et al. [14] investigate why special security controls are necessary for AI and conclude that extending the ISO/IEC 27-family is an effective strategy for AIA compliance.

### C. Goal of this Paper – Interplay between ISMS and AIMS

This paper offers an initial approach to address the following gaps:

1. **Clarification of the interface between ISMS and AIMS through AI security controls within ISMS**: A demonstration of the German national ISO 27001-compliant ISMS, specifically the Grundschutz framework of the BSI (Federal Office for Information Security in Germany). Due to its technical detail, this standard is highly illustrative for the purpose of this paper.

2. **National situation in Germany**: Identifying necessary actions for the BSI in regulating AI according to the AIA.

## II. Methodology

The comparison between the target and actual state of the Information and Communications Technology (ICT)

infrastructure as outlined in the ENISA "Multilayer framework for good cybersecurity practices for AI" [12], also known as the FAICP framework, and the existing Grundschutz framework (2023 edition) [15] is carried out in four steps:

1. **Identification of commonly used use cases of classical and generative AI systems**: Various studies exist that explore real-world applications of AI in different industries. The developed regulations are not future-oriented but as practical as possible.

2. **Characteristics and attributes of identified use cases**: What does the ICT infrastructure look like for the respective use cases? This includes IT infrastructure and hardware, AI models, processes, data, and more.

3. **Modeling using existing modules**: Identification of relevant modules from the Grundschutz framework. AI assets that cannot be modeled with existing modules must be covered by newly developed AI-specific modules (target-actual comparison).

4. **Risk analysis**: Defining module content by setting a reasonable scope, identifying threats, and defining requirements. This also includes creating a cross-reference table. Additional AI-security standards are used to identify controls [16, 17, 18, 19, 20, 21, 22, 23, 24, 25].

To revise personnel competencies, the following steps are performed:

1. **Initial review of existing areas and associated personnel groups and their curricula**: This step examines whether existing areas can be meaningfully supplemented with AI content.

2. **Identification of non-integrable AI content**: Analysis of AI content that cannot be effectively incorporated into existing personnel groups and should be distributed to new groups. A target-actual comparison between competency requirements in the EU AI Act and standards like ISO 42006 is conducted with the current scopes of department SZ 12.

III. RESULTS

The comparison reveals that existing modules can be utilized and new AI-specific modules are necessary to model the ICT infrastructure effectively. The outcome: substantively, the BSI has established a comprehensive basis for implementing the EU AI Act with minimal effort and, as the first regulatory body in the EU, adopting the ISO/IEC 42x series as a national standard. Depending on the application, existing modules such as CON.2 Data Protection, CON.3 Data Backup Concept, APP.4.4 Kubernetes, OPS.2.2 Cloud Use, OPS.2.3 Outsourcing Use, or APP.6 General Software can be used for AI system modeling. Additionally, the introduction of the following four AI modules is recommended to supplement the Grundschutz framework:

- **AI Cyber Governance Module**: This module describes the requirements for the secure implementation and use of AI services. It targets all institutions that already use or intend to develop and deploy such services. The requirements pertain to essential aspects of AI governance, compliance, and monitoring, contributing to the identification and mitigation of potential threats to operations. This module aims to link the EU AI Act's requirements with the Grundschutz framework, ensuring risks remain manageable for institutions. Governance within AIMS should also consider transparency dimensions that are beyond the scope of the CISO's purview.

- **Data Module**: This module addresses the data used in AI applications, including training data utilized for initial model setup and metadata such as source information, quality metrics, and usage logs. Central considerations include data protection and privacy. Sensitive data in AI datasets must be secured per legal requirements to protect user privacy. Robust security measures are necessary to guard AI data against unauthorized access, loss, or manipulation, ensuring data integrity. The provenance and usage of AI data should be documented to build user and stakeholder trust. Continuous monitoring is required to maintain the effectiveness and security of AI applications, as data and threats evolve over time. An AIMS should ensure that data quality, timeliness, provenance, ethical standards, and relevance are maintained.

- **AI Model Module**: This module focuses on AI models, which are characterized by their ability to create and apply mathematically computable models based on data and adapt these models using algorithms. AI uses data and the patterns they contain as learning foundations and, potentially, for generating its own data. The use of AI should be assessed for risks related to information security, manipulation, and environmental impact, following the requirements detailed in this module. An AIMS should build on this to include aspects such as transparency and traceability.

- **AI Platform Module**: This module outlines the requirements for the secure setup and operation of AI services (AI Function-as-a-Service). It supports institutions in the planning, execution, and control of the entire AI system lifecycle, covering both technical and organizational aspects of information security.

With the Grundschutz framework edition 2023 and these four new modules, a broad range of AI systems can be effectively modeled. The newly developed AI modules emphasize the information security aspects of trustworthy AI (dimension: security). Both product and organizational perspectives are considered, analogous to the FAICP framework, covering the entire AI lifecycle from the user's standpoint. These modules complement the C5 and AIC4 catalogs, which are focused on the provider perspective. Grouping these four AI modules into a dedicated module set is logical as they are thematically linked and address specific AI security aspects such as governance, data management, model management, and platform operation. This structured grouping facilitates the consistent handling of complex and interdependent security requirements for AI systems, supports the implementation of the EU AI Act, and simplifies future extensions for emerging technologies or use cases.

## IV. Discussion

The Member States of the European Union are obligated under the EU AI Act to designate national competent authorities by August 2025 to oversee the implementation of the requirements set forth in the AI regulation [26]. The tasks of national oversight of AI essentially encompass three main areas:

- The authority is responsible for appointing independent auditing bodies tasked with the evaluation and control of high-risk AI systems.
- The authority is also in charge of market surveillance, serving as a point of contact for AI providers who identify flaws in their systems.
- The authority should promote and foster innovation and competition.

Article 70 of the EU AI Act establishes the framework for the national competent authorities of the Member States, which are responsible for the implementation and supervision of the regulation [1]. It is conceivable that multiple institutions could jointly assume these responsibilities. In Germany, potential candidates for these tasks include the Federal Office for Information Security (BSI), the Federal Network Agency (BNetzA), data protection authorities, or a newly established agency [27]. So far, no political decision has been made, and there are numerous arguments for each option.

In the past, the BSI has already published studies, the AI criteria catalog AIC4, and various guidelines for the secure implementation of AI. Even if the BSI is not assigned an official role under the AI Act due to political decisions, it is certain that the agency will continue to play an important role in AI regulation. The BSI is actively establishing itself in current regulatory initiatives. An example of this is the working group CEN-CENELEC JTC 21. Furthermore, a horizontal standard for high-risk and so-called GPAI AI systems and their use cases is being developed to operationalize and certify the requirements of the EU AI Act for these categories.

Whether or not the BSI will be chosen as the supervisory authority under the EU AI Act, a measure emerges for the agency from the aforementioned points: the introduction of a new standard, BSI 200-5 (KIMS), which would adopt the essential structure of ISO/IEC 42001, thereby creating a specification of the European AI regulation based on ISO/IEC 42001. While the interfaces between ISMS and AIMS are not clearly defined in the ISO standards, the BSI could delineate these interfaces by introducing new AI modules and BSI 200-5, avoiding content gaps. This interface would involve extending the ISMS to include AI-specific controls that are part of the Grundschutz framework or ISO/IEC 27001 certification.

A significant challenge is also addressed: by nationalizing ISO/IEC 42001 through BSI 200-5, operationalized dimensions of trustworthy AI can be introduced, and technical controls can be made more concrete. The new BSI standard 200-5 could serve as a certification basis for an AIMS in the long term.

In this context, new personnel certifications would be necessary, such as an ISO/IEC 42001 Lead Auditor based on BSI 200-5. The BSI differentiates between the scope of competency assessment and certification of individuals across three areas: competency assessments without individual certification, certification for conducting examinations without aiming for (system) certification, and certification for conducting audits aimed at (system) certification.

In conclusion, the BSI could play a pivotal role in implementing the EU AI Act in Germany by intelligently using and expanding existing standards and processes. By introducing a new BSI standard 200-5, the BSI could create a bridge between international standards, such as ISO/IEC 42001, and the specific requirements of the EU AI Act. This approach would not only create synergies and reduce complexity but also enhance the competitiveness and innovation potential of the German AI landscape. Tailored adjustments to existing certification and competency assessment systems would ensure that the necessary skills and expertise for securely handling AI are built and continuously developed in Germany. Assuming that the BSI develops horizontal CC/evaluation criteria to meet the EU AI Act requirements, a specialized AI deepening for CC evaluators would be advantageous. Such specialized knowledge of the CC evaluator related to a product type, technology, or site type, depending on the area of application, is desired by the BSI. For example, there are already two specialization directions in certain technology areas: Technical Domain "Smartcards and Similar Devices" and Technical Domain "Hardware Devices with Security Boxes." A technical domain for "High-Risk AI Systems" would make sense in the context of an evaluation basis. Alternatively, or in addition, a new group of individuals (Trusted AI Evaluators for High-Risk Systems) could be introduced. It should be noted that these groups do not only focus on the security objectives of AI in terms of security. In parallel with the high-risk category in the AI Act, safety and ethical aspects are prioritized.

From a consulting perspective ("Grundschutz-Berater"), supplementing existing personnel groups is sufficient. The personnel groups of penetration testers and incident experts focus on the trinity of AI and cybersecurity. For GS practitioners and GS consultants, supplements regarding new AI laws and standards are important. Once the BSI introduces the 200-5 standard as the national basis for implementing an AI management system, audit and audit team leaders for certification, analogous to ISO 42006, will be required.

This national German solution described in this chapter could also be adapted to other European frameworks that are based on ISO/IEC 27001.

## V. Conclusion and Future Direction

A comparison between the Grundschutz framework 2023 edition and the ICT infrastructure of the FAICP framework shows that four additional AI modules are necessary to address security risks comprehensively. These modules would be part of the BSI IT Grundschutz certification in the future. With the AI modules, the interfaces between ISMS and AIMS are clearly defined for the first time. The new BSI standard 200-5 would integrate ISO/IEC 42001 and 42006 and adapt them to meet the EU AI Act requirements. Correspondingly, BSI's personnel competencies and certifications should be adjusted. Through a compatible approach with the international standard ISO/IEC 27001, other European countries could similarly adapt their frameworks following the recommendations of this paper.

It is essential to eliminate ambiguities concerning the obligatory nature of organizational security controls under Article 15 of the AIA. Additionally, there is a need to specify the operationalized dimensions of trustworthy AI and the controls in the national standard 200-5 to introduce a European AIMS compliant with the AIA. The operationalization proposals provided by Fraunhofer IAIS can be used as a basis for this approach [28]. Harmonized standards that may be published by the EU Commission and its institutions in the future should be taken into account. The Federal Office for Information Security (BSI) is currently developing the new "Grundschutz++" framework, and it could be beneficial to incorporate the new AI modules and the new AI standard as integral components. Fundamentally, however, it remains to be seen what role the BSI will play in the future regulation of AI. Although regulatory oversight by the BSI appears reasonable from a technical perspective, it ultimately remains a political decision.